\documentclass[11pt]{article}

\usepackage[utf8]{inputenc}
\usepackage[T1]{fontenc}
\usepackage[margin=2cm]{geometry}
\usepackage{amsmath,amssymb}
\usepackage{booktabs}
\usepackage{xcolor}
\usepackage[numbers,sort&compress]{natbib}
\usepackage{float}
\usepackage{graphicx}
\usepackage{algorithm}
\usepackage{algpseudocode}
\usepackage[colorlinks=true,linkcolor=blue!50!black,citecolor=blue!50!black,urlcolor=blue!50!black]{hyperref}
\title{Kairos: Toward a Byte-Level Foundation Model for Every Modality\\
\large A Pyramidal Codec and Denoising Diffusion over Bidirectional Delta/Window-Attention\\
\large Mixture-of-Experts -- Does It Converge? Sanity Checks at Tiny Scale}

\author{Fabien Furfaro\thanks{Code and configurations:
\url{https://github.com/fabienfrfr/Kairos}}}
\date{September 2026}

\usepackage{titling}
\posttitle{\par\end{center}\vspace{-0.6cm}}
\preauthor{\begin{center}\large}
\postauthor{\par\end{center}\vspace{-0.9cm}}

\begin{document}
\maketitle

\begin{abstract}
\textbf{Background:} Small language models trained on a single local machine leave
little room for architectural trial and error. Kairos combines bidirectional gated-DeltaNet linear attention and sliding-window attention
over shared projections (LiZAttention2), block-wise Attention-Residual aggregation, a
sparse Mixture-of-Experts (MoE) feed-forward layer and a multi-scale byte-level codec,
trained with a uniform-noise denoising objective over one byte space shared by text, image,
audio, video, lidar and control streams; this report asks whether the combination optimizes
at all before asking whether it is useful.
\textbf{Methods:} We give code-derived algorithms for each component, including a
cross-session recurrent-state memory gate, then run convergence sanity checks: overfitting
a single 440-byte sentence at 1--15M parameters, text only, with three seeds for codec and
attention choices and single-run sweeps for seven further factors.
\textbf{Results:} The two attention variants tested at 3\,000 steps (LiZAttention2 and
window attention alone) reach a minimum loss near $10^{-3}$. After 600 steps, the mean
minimum loss ranges from 0.018 (patch codec) to 0.354 (depthwise convolutional codec) with
an identical ranking on three seeds. An apparent advantage of shared over separate
attention projections closes with more steps or more diverse data and reverses when the
projections are frozen: it is a convergence-speed effect, not a capacity effect.
\textbf{Conclusions:} These checks suggest that the implementation optimizes; they say
nothing about generalization, multimodal training or comparison with existing models, which
a planned pretraining run is meant to address. The Discussion sizes a 200M-total/25M-active
target analytically; none of its throughput figures is measured.
\end{abstract}

\begin{figure}[H]
    \centering
    \includegraphics[width=\linewidth]{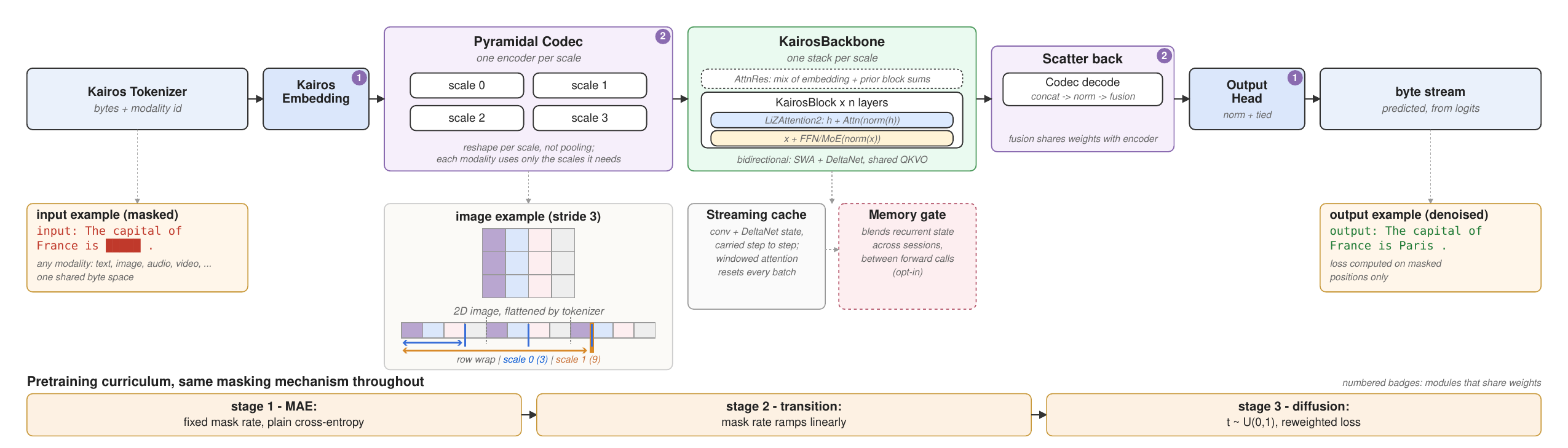}
    \caption{Kairos data and training flow, left to right. The tokenizer (\S\ref{sec:tokenizer})
    flattens every modality to one byte stream; the image example shows a 2D image flattened row by
    row (its \texttt{<ENDLINE>} markers are omitted). Each KairosBlock is a standard
    pre-norm attention+FFN residual pair; AttnRes only re-mixes what feeds into it. Training (bottom)
    uses one masking mechanism across three curriculum stages (MAE $\to$ transition $\to$ diffusion),
    illustrated with a masked-input/denoised-output example.}
    \label{fig:architecture_overview}
\end{figure}

\section{Introduction}

Small, edge-deployable language models
(MobileLLM~\citep{liu2024mobilellm}, SmolLM2~\citep{allal2025smollm2},
Gemma~3~\citep{gemma3}) are routinely trained on far more tokens than the
Chinchilla-optimal ratio~\citep{hoffmann2022chinchilla}, on the premise that
inference cost dominates total cost of ownership. That premise says nothing
about training cost when the training hardware is itself constrained, the
setting of this report. Three architectural levers target efficiency in
largely separate literatures: (i) sub-quadratic sequence mixing, through
local windowed attention~\citep{beltagy2020longformer,jiang2023mistral} and
linear state-compressing mixers derived from state-space models
(Mamba~\citep{gu2023mamba}) and the delta
rule~\citep{yang2024deltanet,yang2024gateddeltanet}; (ii) sparse capacity
scaling through MoE
routing~\citep{shazeer2017outrageously,fedus2021switch,dai2024deepseekmoe,deepseekv3};
(iii) non-autoregressive discrete diffusion
training~\citep{austin2021d3pm,lou2023sedd,sahoo2024mdlm,nie2025llada}, which
generalizes masked language modeling~\citep{devlin2018bert}. A fourth,
orthogonal choice removes the tokenizer: byte-level
inputs~\citep{xue2022byt5,yu2023megabyte,pagnoni2024blt} give one vocabulary
that can cover several modalities, an idea we had explored at smaller scale for two
modalities~\citep{furfaro2024pixelbytes} and revise substantially here for six.

Each lever is usually reported on its own. Combining them in one model small enough to
train on a single local machine raises a practical question before any question of
quality: does the combination optimize at all? Hybrid attention, sparse routing, a
multi-scale codec and a corruption-based objective each add moving parts that could
prevent optimization, and a memorization test is the cheapest way to check.
\textbf{Research question:} can a Kairos network drive the denoising loss of a fixed byte
string to near zero, and which design choices change how fast?

\textbf{In this paper}, we specify the Kairos architecture from its implementation
(Figure~\ref{fig:architecture_overview}) and report convergence sanity checks at
1--15M parameters. These checks are necessary and not sufficient: they do not measure
generalization. Held-out training and a comparison with Pythia-14M~\citep{biderman2023pythia}
are planned and not reported (\S\ref{sec:future}).

\textbf{Our contributions are:}
\begin{enumerate}
  \item The Kairos architecture (Figure~\ref{fig:architecture_overview}), specified from its implementation:
        the multimodal byte tokenizer (\S\ref{sec:tokenizer}), PyramidalCodec (Algorithm~\ref{alg:codec}), LiZAttention2
        (Algorithm~\ref{alg:liz}), Block-AttnRes
        (Algorithm~\ref{alg:attnres}), sparse MoE (\S\ref{sec:moe}), a
        cross-session memory gate (Algorithm~\ref{alg:memgate}) and the
        denoising objective with its curriculum
        (Algorithm~\ref{alg:diffusion}), each adapted from a specific prior
        method or, for the codec and the memory gate, combining existing
        ideas.
  \item A convergence study at tiny scale (\S\ref{sec:ablations}) with 3-seed
        statistics for codec and attention choices and single-run sweeps for
        seven further factors.
  \item A worked correlation-versus-causation case
        (\S\ref{sec:qkv-sharing}): a clear single-run effect that does not
        survive three controls, and a recommendation to report
        steps-to-threshold rather than loss at a fixed step.
\end{enumerate}
The Discussion also sizes an untrained 200M-total/25M-active target with a roofline argument
(\S\ref{sec:sizing}); it is analytical and not measured.

\section{Related Work}

\paragraph{Efficient attention.} Full self-attention~\citep{vaswani2017attention}
is quadratic in sequence length. The delta
rule~\citep{yang2024deltanet,yang2024gateddeltanet} improves plain linear
attention with an error-correcting state update; Kairos runs it
\emph{bidirectionally} (a forward pass and a time-reversed pass,
concatenated) to match the context a denoising objective needs.
Sliding-window attention~\citep{beltagy2020longformer,jiang2023mistral} is
likewise bidirectional here, with rotary position
embeddings~\citep{su2021roformer}; the code supports grouped-query
attention~\citep{ainslie2023gqa}, which the configurations reported here do
not exercise. Coupling a delta-rule branch and a window-attention branch
through shared projections follows TPTT~\citep{furfaro2025tptt}.

\paragraph{Byte-level, hierarchical and multimodal models.} ByT5 removes the
subword tokenizer~\citep{xue2022byt5}; Kairos' tokenizer subclasses it and
extends it to other modalities. MegaByte models bytes with a global model over
fixed-size patches and a local model inside each
patch~\citep{yu2023megabyte}; the Byte Latent Transformer forms patches
dynamically from next-byte entropy~\citep{pagnoni2024blt}; Hourglass
Transformers shorten and re-expand the sequence inside one
stack~\citep{nawrot2022hourglass}; Perceiver~IO reads arbitrary input arrays,
including raw bytes, through cross-attention into a fixed latent
array~\citep{jaegle2022perceiverio}. Kairos' PyramidalCodec instead keeps
$S$ parallel views of one sequence at geometrically growing patch sizes,
runs a backbone on each, and fuses them after upsampling. 

\paragraph{Sparse MoE.} MoE
architectures~\citep{shazeer2017outrageously,fedus2021switch,dai2024deepseekmoe,deepseekv3}
decouple capacity from per-token compute. \texttt{KairosMoE} subclasses
DeepSeek-V3's routed-expert layer~\citep{deepseekv3} (sigmoid routing scores,
shared experts, and load balancing by a per-expert selection bias rather than
an auxiliary loss), and \texttt{KairosFFN}, used when MoE is disabled,
subclasses Qwen2's dense MLP~\citep{qwen2techreport}. Only the expert-weight
initialization is patched, because the upstream implementation used leaves
those weights uninitialized.

\paragraph{Cross-layer residual aggregation.} Block-AttnRes replaces the
fixed, unit-weight residual accumulation standard since
ResNet~\citep{he2015resnet} with the learned, block-windowed softmax
attention over earlier layer outputs of \citet{kimiteam2026attnres}, used
here as published (\S\ref{sec:attnres}).

\paragraph{Discrete diffusion language modeling.} Kairos' corruption is
closest to the uniform-replacement transition of D3PM~\citep{austin2021d3pm}:
corrupted positions are filled with tokens drawn uniformly from the
vocabulary, and the model is not told which positions were corrupted, unlike
the absorbing mask state of MDLM~\citep{sahoo2024mdlm}. The loss is weighted
by $1/p$ in the manner of MDLM, with a clip, and trained with
self-conditioning~\citep{chen2022analogbits} and a three-stage curriculum
(\S\ref{sec:objective}).

\paragraph{Recurrent memory.} Transformer-XL carries hidden states across
segments~\citep{dai2019transformerxl}, Titans learn a test-time
memory~\citep{behrouz2025titans}, and TPTT retrofits such memory into
pretrained transformers~\citep{furfaro2025tptt}. Kairos' gate is simpler: it
seeds the recurrent state of each DeltaNet layer for a new sequence from a
learned blend of states left by earlier sequences (\S\ref{sec:memgate}).

\paragraph{Our positioning.} Kairos assembles components from the works above and adds
two elements: a pyramidal codec that runs the backbone on several patch scales, and a
cross-session memory gate. We did not perform a systematic literature search and claim no novelty beyond this combination. The aim of this report is to check that the
combination trains, not that it outperforms.

\section{Model}

Figure~\ref{fig:architecture_overview} gives the data flow. The subsections follow it from
input bytes to output logits, then the training objective.

\subsection{Multimodal byte tokenizer}
\label{sec:tokenizer}

The tokenizer subclasses the ByT5 tokenizer~\citep{xue2022byt5}. Text is UTF-8 and every byte is
one id, with the same byte-to-id offset in all modalities, so one vocabulary of 290 entries (3 ByT5
special ids, 256 byte values, 31 added markers) serves every modality. A sequence is a list of
typed segments, each wrapped in modality delimiters (for example \texttt{<IMG>}\ldots\texttt{</IMG>})
and optionally in channel delimiters (R, G, B, left, right). Encoding returns three aligned
streams: the ids, a modality id per position, which drives the scale router of the codec, and an
\emph{octet-family} id per position, which names the byte plane a byte belongs to. Families live
in a separate stream rather than in the vocabulary, so one byte value can carry different meanings
in different planes (18 family ids, family 0 being text and markers).

Non-text values are serialized without a header (Table~\ref{tab:tokenizer}). A float
$x\in[\ell,u]$ becomes $n$ big-endian byte planes, as the digits of a place-value number (Eq.~\ref{eq:quant}):
\begin{equation}
q=\Big\lfloor \text{clip}\Big(\tfrac{x-\ell}{u-\ell},0,1\Big)\,(256^{n}-1)\Big\rfloor,\qquad
b_k=\Big\lfloor q/256^{\,n-1-k}\Big\rfloor \bmod 256,
\label{eq:quant}
\end{equation}
with $n=2$ by default and fixed ranges ($[-1,1]$ for audio and control signals, $\pm100$ for lidar
coordinates, $[0,1]$ for intensity), so the encoding needs no per-sample header. Optional
downsampling before quantization (block mean for pixels, decimation by 4 for audio and signals by
default, point stride for lidar) makes it lossy. Decoders recover shapes by counting markers and
raise an error on a malformed stream, so a generated stream can be checked.

\begin{table}[t]
\centering
\small
\begin{tabular}{p{1.7cm}p{6.2cm}p{4.3cm}r}
\toprule
Modality & Byte serialization & Markers & Families \\
\midrule
Text & UTF-8 bytes & none & 1 \\
Image & interleaved R, G, B bytes, row by row & \texttt{<ENDLINE>} per row & 3 \\
Video & image frames in sequence & \texttt{<ENDLINE>}, \texttt{<ENDFRAME>} & 3 (shared) \\
Audio & 2-byte planes of decimated samples & \texttt{<TICK>} every 16{,}000 samples & 2 \\
Lidar & $(x,y,z,\text{intensity})$, 2 bytes each & \texttt{<PTSEP>} every 32 points & 8 \\
Control & action and state, 2 bytes each, interleaved & \texttt{<OBS>} for state-only & 4 \\
\bottomrule
\end{tabular}
\caption{Every modality reduces to bytes plus at most three kinds of side information: a
delimiter, a structural marker and a family id (\texttt{kairos/tokenizer.py}). The control
serialization follows the action--state encoding of~\citet{furfaro2024pixelbytes}.}
\label{tab:tokenizer}
\end{table}

\subsection{Embedding and PyramidalCodec}

Each token id and its octet-family id are embedded and fused by a linear layer (\S\ref{sec:tokenizer});
the output head, tied to the token embedding, predicts both the byte and its family.

The PyramidalCodec builds $S$ parallel views of the sequence with patch sizes
$p_s=\text{stride}^s$ (notebook config: stride $3$, $S=4$, so $p=3,9,27,81$). A shared
rank-$r$ bottleneck mixer
($\text{Conv1d}_{d\to r}\to\text{GELU}\to\text{Conv1d}_{r\to d}$,
$r=\max(8,d/4)$, cost $O(rd)$ against $O(d^2)$ for a dense $1{\times}1$
convolution) supplies cross-channel context once; each scale then applies a
grouped convolution with kernel and stride $p_s$ and its own mixer
(Algorithm~\ref{alg:codec}). The group count is the one knob that trades
compute for expressiveness: \texttt{channels\_per\_group}$=1$ (default) is
depthwise and cheapest, and raising it toward $d$ approaches a dense
convolution. \texttt{mode="patch"} replaces the convolution and mixer chain by
one $\text{Linear}(p_s d\to d)$ per scale, decoded with the transposed weight
(tied encode and decode). Each scale runs through its own backbone, or through
one shared backbone (about $75\%$ fewer backbone parameters for $S=4$). A
scale router lets each modality read only a subset of scales (notebook config: text
$\to$ scales 1--2, control 1, image 2--3, lidar 2, audio 3, video 3--4,
metadata 4); positions of modalities not routed to a scale bypass its
backbone. Decoding mirrors encoding with grouped transposed convolutions and
per-scale mixers, then concatenates the scales, applies RMSNorm and fuses them
with a linear layer.

\begin{algorithm}[t]
\caption{PyramidalCodec, \texttt{mode="conv"}: encode, per-scale backbone, decode}
\label{alg:codec}
\begin{algorithmic}[1]
\Require $x\in\mathbb{R}^{B\times L\times d}$; patch sizes $p_1,\ldots,p_S$; backbones $f_1,\ldots,f_S$
\State $h\gets\text{Mixer}_{\text{pre}}(x)$ \Comment{shared channel mixing, computed once}
\For{$s=1$ \textbf{to} $S$}
  \State $h_s\gets\text{GroupedConv1d}\big(\text{PadToMultiple}(h,p_s);\ \text{kernel}=\text{stride}=p_s\big)$
  \State $z_s\gets\text{Mixer}_s(h_s)$ \Comment{$\lceil L/p_s\rceil$ positions}
  \State $\tilde z_s\gets f_s(z_s)$ \Comment{active positions only, chosen by the scale router}
  \State $u_s\gets\text{Trim}_L\big(\text{Mixer}'_s(\text{ConvTranspose1d}(\tilde z_s))\big)$
\EndFor
\State \Return $\text{Linear}_{Sd\to d}\big(\text{RMSNorm}(\text{concat}(u_1,\ldots,u_S))\big)$
\end{algorithmic}
\end{algorithm}

\subsection{LiZAttention2: shared-projection bidirectional hybrid attention}
\label{sec:liz}

Each layer runs a bidirectional sliding-window branch (mask $|q-k|\le w$, RoPE)
and a bidirectional gated-DeltaNet branch over the same normalized input. The
DeltaNet branch's Q/K/V and output projections are deleted and replaced by
references to the window branch's, so both mixers read the same linear
projections of $x$ (\texttt{liz2\_share\_qkv=False} restores independent
ones; \S\ref{sec:qkv-sharing}). The DeltaNet branch keeps its own value
expansion (factor 2), depthwise causal convolution, decay and write-strength
gates, output gate, and a merge of its forward and time-reversed passes
(Algorithm~\ref{alg:liz}). The two branch outputs are concatenated,
normalized with RMSNorm~\citep{zhang2019rmsnorm} and projected back to $d$.

\begin{algorithm}[t]
\caption{LiZAttention2 forward (one layer)}
\label{alg:liz}
\begin{algorithmic}[1]
\Require $x\in\mathbb{R}^{B\times L\times d}$; shared projections $W_q,W_k,W_v,W_o$
\State $o^{\text{swa}}\gets W_o\,\text{BidirWindowAttn}\big(\text{RoPE}(W_qx),\text{RoPE}(W_kx),W_vx;\,w\big)$
\For{$\text{dir}\in\{\text{fwd},\text{bwd}\}$}
  \State $x'\gets x$ if fwd, else $\text{flip}(x)$ \Comment{flip along the sequence axis}
  \State $(q,k,v)\gets\text{SiLU}\,\text{DepthwiseCausalConv}\big([W_qx';\,W_kx';\,E_vW_vx']\big)$ \Comment{$E_v$: value expansion}
  \State $\beta\gets\sigma(W_\beta x')$;\quad $g\gets-\exp(A_{\log})\,\text{softplus}(W_ax'+b_{dt})$ \Comment{per head}
  \State $o^{\text{dir}}\gets\text{GatedDeltaRule}(q,k,v;\,g,\beta)\odot\text{SiLU}(W_gx')$ \Comment{$q,k$ L2-normalized}
\EndFor
\State $o^{\Delta}\gets W_o\,W_{\text{lr}}\,\text{RMSNorm}\big([o^{\text{fwd}};\,\text{flip}(o^{\text{bwd}})]\big)$
\State \Return $W_{\text{mix}}\,\text{RMSNorm}\big([o^{\text{swa}};\,o^{\Delta}]\big)$ \Comment{$2d\to d$}
\end{algorithmic}
\end{algorithm}

\subsection{Block-AttnRes aggregation}
\label{sec:attnres}

Instead of the plain residual stream $x_{i+1}=x_i+f_i(x_i)$, each layer's
\emph{input} is a learned, softmax-weighted mix of the embedding, the
finalized block sums of earlier layers, and a running partial sum of the
current block~\citep{kimiteam2026attnres} (Algorithm~\ref{alg:attnres}); the
sublayers of $f_i$ still use ordinary residual connections around their own
input. The weight vector $w$ of each aggregator is initialized to zero, so the
initial mix is uniform. With block size $1$ each earlier layer output is a
separate source, a densely connected pattern~\citep{huang2016densenet} with learned
softmax weights; with block size $n_{\text{layers}}$ the sources reduce to the
embedding and one running sum, and larger sizes change nothing. The notebook config uses
block size $4$ over $4$ layers; the ablation networks have one layer, where
the aggregator is a structural no-op.

\begin{algorithm}[t]
\caption{Block-AttnRes backbone forward}
\label{alg:attnres}
\begin{algorithmic}[1]
\Require embedding $x_0$; layers $f_1,\dots,f_L$; block size $S$; learned $w_i\in\mathbb{R}^d$ per aggregator
\State $\text{completed}\gets[\,]$;\ $\text{partial}\gets\text{None}$;\ $\text{count}\gets0$
\For{$i=1,\dots,L$}
  \State $\text{sources}\gets[x_0]\cup\text{completed}\cup(\{\text{partial}\}\text{ if not None})$
  \State $V\gets\text{stack}(\text{sources})$;\quad $K\gets\text{RMSNorm}(V)$
  \State $h\gets\sum_{\text{sources}}\text{softmax}(w_i^{\top}K)\cdot V$ \Comment{softmax over sources}
  \State $x_i\gets f_i(h)$
  \State $\text{partial}\gets x_i$ if partial is None else $\text{partial}+x_i$;\quad $\text{count}\mathrel{+}=1$
  \If{$\text{count}=S$}
    \State $\text{completed}.\text{append}(\text{partial})$;\ $\text{partial}\gets\text{None}$;\ $\text{count}\gets0$
  \EndIf
\EndFor
\State \Return $\text{RMSNorm}(x_L)$
\end{algorithmic}
\end{algorithm}

\subsection{Sparse Mixture-of-Experts}
\label{sec:moe}

Each token is routed to $k$ of $E$ routed experts plus one or more always-on
shared experts. Balance is maintained by nudging a per-expert selection bias
toward equal load (DeepSeek-V3 style, no auxiliary loss by default). The
active-parameter count used throughout is computed directly:
\begin{equation}
N_{\text{act}}=N_{\text{total}}-N_{\text{expert}}+N_{\text{expert}}\frac{k}{E},
\label{eq:active}
\end{equation}
where $N_{\text{expert}}$ sums the parameters under modules whose name
contains \texttt{".experts."}; shared experts, the router, the attention and
DeltaNet path and the codec count as fully active
(\texttt{count\_active\_parameters}).

\subsection{KairosMemoryGate: cross-session recurrent-state memory}
\label{sec:memgate}

With \texttt{use\_memory\_gate=True}, the recurrent state of every DeltaNet
layer of every scale is seeded for a new sequence not from zero but from a
gate over a bank of states left by earlier sequences (the preceding batch's
final state, or an externally supplied bank). One gate module is shared by all
scales and layers. It is a single-head cross-attention over a rank-$r'$
bottleneck, $r'=\max(8,\text{round}(\sqrt{D}/4))$, where $D$ is the flattened
DeltaNet state size ($D=2048$ and $r'=11$ for the notebook config). The query is the
projection of the (zero) current state; the keys and values are that same
projection followed by the projected bank, so the gate can attend to the
bank or fall back on the current state (Algorithm~\ref{alg:memgate}). It is
applied to the forward pass only; the time-reversed pass starts from zero.

\begin{algorithm}[t]
\caption{Memory gate, applied per (scale, DeltaNet layer)}
\label{alg:memgate}
\begin{algorithmic}[1]
\Require zero state $s_0\in\mathbb{R}^{B\times D}$; bank $M\in\mathbb{R}^{n\times D}$ of flattened prior states
\If{$n=0$} \State \Return $s_0$ \EndIf
\If{$n=1$} \State \Return $M_1$ broadcast to $B$ rows \EndIf
\State $q\gets\text{Linear}_{D\to r'}(s_0)$;\quad $K\gets[\,q;\ \text{Linear}_{D\to r'}(M)\,]$
\State $o\gets\text{MultiheadAttention}_{h=1}(q,K,K)$
\State \Return $\text{Linear}_{r'\to D}(o)$
\end{algorithmic}
\end{algorithm}

\subsection{Denoising objective and curriculum}
\label{sec:objective}

Given a clean byte sequence $x_0$, each row draws a rate
$p\sim U(\epsilon,p_{\max})$ and a Bernoulli$(p)$ mask over non-prompt,
non-padding positions. Masked positions are replaced by tokens drawn
uniformly from the vocabulary (and random octet-family ids), giving $x_t$; no
mask indicator is provided. The model predicts $x_0$ at the masked positions
with cross-entropy, weighted per token by $\min\!\big(c,\,1+\alpha(1/p-1)\big)$,
a blend between plain cross-entropy ($\alpha=0$) and full $1/p$ weighting
($\alpha=1$) in the manner of MDLM~\citep{sahoo2024mdlm}, clipped at $c$ to
bound the variance from rows with small $p$. With probability $0.5$ per batch,
self-conditioning~\citep{chen2022analogbits} feeds a detached, no-gradient
first prediction back in as an additive embedding, matching its use in
\texttt{generate()}. The default is $\epsilon=10^{-3}$, $c=3$; an
octet-family cross-entropy term with weight $1$ is added when family ids are
present.

\begin{algorithm}[t]
\caption{Denoising training step (\texttt{compute\_loss})}
\label{alg:diffusion}
\begin{algorithmic}[1]
\Require batch $x_0$, prompt lengths, pad mask; schedule values $(p_{\max}^{(t)},\alpha^{(t)})$; clip $c$
\State $p_b\gets\epsilon+(p_{\max}^{(t)}-\epsilon)\,u_b$,\ $u_b\sim U(0,1)$ \Comment{one rate per row}
\State $m\gets\text{Bernoulli}(p_b)$ restricted to non-prompt, non-pad positions
\State $x_t\gets x_0$ with $x_t[m]\gets$ tokens drawn uniformly from the vocabulary
\If{$u<0.5$, $u\sim U(0,1)$} $\hat y\gets\text{stopgrad}(\text{model}(x_t))$; add $\text{softmax}(\hat y)E$ to the embeddings \EndIf
\State $\text{logits}\gets\text{model}(x_t,\ \text{logits\_mask}=m)$ \Comment{project masked positions only}
\State $\ell_i\gets\text{CE}(\text{logits}_i,x_{0,i})\cdot\min\big(c,\,1+\alpha^{(t)}(1/p_{b(i)}-1)\big)$
\State \Return $\text{mean}_i\,\ell_i$ \Comment{$+$ octet-family CE when family ids are present}
\end{algorithmic}
\end{algorithm}

The schedule $(p_{\max}^{(t)},\alpha^{(t)})$ is a pure function of the global
step (\texttt{stage\_mask\_schedule}), so resuming from a checkpoint depends
only on the persisted step count. It is flat at $(0.3,\,0)$ in the MAE stage,
ramps linearly to $(1.0,\,1)$ over the transition stage, then stays flat in
the diffusion stage. The staging was introduced after an informal observation
of optimization instability when sampling the full masking range from a random
initialization; that observation is not quantified here, and no run without
the curriculum is reported.

\section{Experimental Protocol}

This section fixes the three model scales, then the overfitting protocol behind every
measured number in \S\ref{sec:ablations}.

\subsection{Three scales, kept apart}

The codebase is used at three scales that this report does not conflate
(Table~\ref{tab:configs}). The \textbf{ablation networks} of
\S\ref{sec:ablations} use the library defaults of
\texttt{configs/ablations/baseline.yaml} (1 layer, dense feed-forward). The
\textbf{notebook configuration} is the default of the project's notebook
(\texttt{notebook/kairos\_pretraining.py}): a $\sim$14--15M-parameter network
with a 7-expert top-1 MoE, used for pipeline sanity checks. The
\textbf{target} is a 200M-total/25M-active design point used in
\S\ref{sec:sizing} purely to size a future run; it has never been
instantiated. Because dense parameters (attention, codec, embeddings) are
always active, $N_{\text{act}}=N_{\text{total}}/8+\tfrac78 N_{\text{dense}}$
for $E/k=8$, and 25M is the lower bound reached when all parameters are
routed experts.

\begin{table}[t]
\centering
\small
\begin{tabular}{@{}lp{3.4cm}p{3.9cm}p{2.6cm}@{}}
\toprule
 & Ablation baseline & Notebook config & Target (sizing only) \\
\midrule
$d_{\text{model}}$; heads (head dim) & 64; 4 (16) & 64; 4 (16) & -- \\
Backbone layers & 1 & 4 & -- \\
Codec stride; scales $S$ & 5; 4 & 3; 4 & -- \\
Codec mode & \texttt{conv}, 1 channel per group & \texttt{conv}, 8 channels per group & -- \\
AttnRes block size & 1 & 4 & -- \\
Feed-forward & dense, $I=2048$ & 7 routed ($k=1$) $+$ 1 shared, $I=544$ & 32 routed ($k=4$) \\
Backbone shared across scales & no & yes & -- \\
Memory gate & off & on & -- \\
$N_{\text{total}}$ & 1.98M & $\sim$14--15M & 200M \\
$N_{\text{act}}$ & $N_{\text{total}}$ & $N_{\text{total}}-2.51$M (Eq.~\ref{eq:active}) & 25M \\
Training data & 16 identical sentences, 600 steps & sanity checks & 30B tokens (budget) \\
\bottomrule
\end{tabular}
\caption{The three scales this report distinguishes. Dashes: the target
design point is specified only by its parameter and token budget. The notebook config's
$2.51$M is derived from its configuration
($4\times7\times3\times64\times544=2.92$M routed-expert parameters, of which
6/7 are inactive); the instantiated count has not been logged. The ablation
networks are far smaller than the notebook config, so ablation findings are not
statements about the notebook config.}
\label{tab:configs}
\end{table}

\subsection{Overfitting protocol}
\label{sec:protocol}

The purpose is a convergence check. A network that cannot drive the loss of a fixed
byte string to near zero has an optimization or implementation defect; success is necessary
and not sufficient for a useful model, since it says nothing about generalization.

All ablations use \texttt{kairos overfit} on the configurations of
\texttt{configs/ablations/} (\texttt{kairos==0.4.1.dev46},
\texttt{torch==2.14.0}), on CPU with eager attention and the PyTorch fallback
DeltaNet path. The data are 16 identical copies of one 440-byte string (the
sentence ``the quick brown fox jumps over the lazy dog'' repeated ten times),
truncated to 128 tokens, i.e.\ about $2.9$ repetitions per sequence; the
63-sentence control uses 63 distinct short sentences. Training uses batch
size 4, 600 steps unless noted, $p_{\max}=0.3$ and no reweighting; the
attention window is the library default $w=128$, which exceeds the 26
positions of the finest scale, so the window branch attends globally. Every
step draws a fresh corruption, and \emph{min.\ loss} is the minimum over the
run of the per-step training loss, an extreme value of a noisy per-batch
quantity. Training and evaluation data coincide, so the study measures
memorization speed, not generalization. Runs are not bit-reproducible even at
a fixed seed: the baseline at seed 0 gave $0.227$ in one run and $0.31$ in
another. Seeds are 0, 1, 2 where stated; standard deviations are sample
standard deviations ($n-1$).

\section{Results}

All results are convergence sanity checks (\S\ref{sec:protocol}) on the ablation networks of
Table~\ref{tab:configs}: the training data are the evaluation data, so they measure
memorization speed, not generalization. \S\ref{sec:ablations} compares codecs and attention
variants, \S\ref{sec:qkv-sharing} reports a single-run effect that did not survive controls,
and the last subsection covers the curriculum and the memory gate.

\subsection{Convergence of the tiny-scale networks}
\label{sec:ablations}

Table~\ref{tab:seeds} reports the codec and attention choices under three
seeds, on the setup of \S\ref{sec:protocol}. Codec ranking is identical on
every seed: \texttt{patch} $<$ dense \texttt{conv} $<$ \texttt{conv} with 8
channels per group $<$ depthwise \texttt{conv} (\texttt{liz2}
baseline), from $0.018$ to $0.354$. The hybrid attention beats SWA alone on
every seed ($0.227$, $0.258$, $0.576$ against $0.596$, $0.728$, $0.798$),
although the mean $\pm$ sd ranges overlap. The notebook config uses \texttt{conv} with 8
channels per group, a compromise between the depthwise default and the dense
variants.

\begin{table}[t]
\centering
\small
\begin{tabular}{lrrrrr}
\toprule
Configuration & Seed 0 & Seed 1 & Seed 2 & Mean $\pm$ sd & Params \\
\midrule
\texttt{liz2}, depthwise codec (baseline) & 0.227 & 0.258 & 0.576 & $0.354 \pm 0.193$ & 1.98M \\
\texttt{vanilla} (SWA only), depthwise codec & 0.596 & 0.728 & 0.798 & $0.707 \pm 0.103$ & 1.81M \\
\texttt{conv} codec, 8 channels per group & 0.106 & 0.045 & 0.075 & $0.075 \pm 0.031$ & 2.50M \\
\texttt{conv} codec, dense (64 channels per group) & 0.037 & 0.024 & 0.024 & $0.028 \pm 0.007$ & 8.10M \\
\texttt{patch} codec (one \texttt{Linear} per scale) & 0.019 & 0.016 & 0.018 & $0.018 \pm 0.002$ & 4.93M \\
\bottomrule
\end{tabular}
\caption{Minimum training loss after 600 steps, three seeds
(\nolinkurl{docs/experiments/ablations_codec_attention.md}, \S1). Depthwise
saves parameters relative to \texttt{patch} (1.98M against 4.93M) at the cost
of slower memorization; 8 channels per group recovers most of the gap for
about half of \texttt{patch}'s parameters (2.50M against 4.93M). The
\texttt{liz2}-versus-\texttt{vanilla} gap is a convergence-speed effect
(\S\ref{sec:qkv-sharing}).}
\label{tab:seeds}
\end{table}

The seed spread ($\pm0.002$ to $\pm0.193$) sets the noise floor for the
single-run sweeps of Table~\ref{tab:sweeps}, so differences of a few
hundredths there are not readable. Monotone trends are consistent with the
expected direction but confounded with the setup: at 128 tokens a sequence
holds about 2.9 verbatim repetitions of the sentence (1.5 at 64, 5.8 at 256),
so sequence length and the number of codec scales change how many copies are
available for in-context recovery of masked bytes, and the loss reflects
copying as much as memorization. A batch of 16 covers the whole training set.
The attention-block-size result is a saturation effect: sizes at or above the
number of layers give the same computation.

\begin{table}[t]
\centering
\small
\begin{tabular}{p{4.6cm}p{9.6cm}}
\toprule
Factor (base configuration) & Min.\ loss by setting (single run) \\
\midrule
Depth $n_{\text{layers}}$ (\texttt{liz2} baseline) & 1: 0.227;\ 2: 0.843;\ 4: 0.368;\ 8$^{\dagger}$: 1.228 \\
MoE routing, 7 experts & top-1: 0.281;\ top-2: 0.272 \\
Reweighting clip $c$ (diffusion regime, \texttt{conv}, 8 channels per group) & $c=3$: 0.139;\ $c=30$: 0.249 \\
AttnRes block size (4 layers) & 4: 0.386;\ 8: 0.386;\ 16: 0.386 \\
Codec scales $S$ & 2: 0.430;\ 3: 0.308;\ 4: 0.227 \\
Sequence length (tokens) & 64: 1.129;\ 128: 0.227;\ 256: 0.035 \\
Batch size & 1: 0.890;\ 4: 0.227;\ 16 (whole set): 0.032 \\
\bottomrule
\end{tabular}
\caption{Single-run sweeps; read as indicative (\S1 of the log gives the
noise floor). $^{\dagger}$300 steps because of compute budget. The default
$c=10$ is omitted: the log has no per-run record for it. Depth does not help
memorization and top-1 and top-2 routing are indistinguishable at this noise
level.}
\label{tab:sweeps}
\end{table}

\subsection{A correlation-versus-causation case: shared versus separate QKV}
\label{sec:qkv-sharing}

Sharing the Q/K/V/O projections between the two attention branches
(\S\ref{sec:liz}) first looked like a clear architectural win: with sharing
disabled, min.\ loss rose from $0.227$--$0.369$ to $0.610$--$0.858$ across
repeated 600-step runs. Three controls were run before accepting this as an
architectural effect (Table~\ref{tab:qkv-sharing}).

\begin{table}[t]
\centering
\small
\begin{tabular}{lrrl}
\toprule
Check & Shared & Separate & Outcome \\
\midrule
600 steps, 16 identical sentences & 0.369 & 0.610 & shared ahead \\
3\,000 steps, same data & 0.0011 & 0.0011 & gap closes \\
63 distinct sentences, 600 steps & 2.269 & 2.285 & gap closes \\
QKV frozen after step 100 & 0.705 & 0.521 & \textbf{gap reverses} \\
\bottomrule
\end{tabular}
\caption{Min.\ loss for shared and separate projections
(\nolinkurl{docs/experiments/ablations_codec_attention.md}, \S7).
The advantage was faster early convergence, not higher capacity: with enough
steps or diverse data both variants reach the same floor.}
\label{tab:qkv-sharing}
\end{table}

The same check applies to \texttt{liz2} against \texttt{vanilla}: at
3\,000 steps \texttt{vanilla} reaches $0.0014$, matching \texttt{liz2}'s
$0.0011$. Computing steps-to-threshold from the saved curves shows that the
shared-versus-separate gap is concentrated in the $0.5$--$0.1$ loss range
($+9$ to $+14\%$ steps for separate projections) and vanishes elsewhere,
which a single comparison at a fixed step count would have hidden. A plausible
mechanism is that shared projections receive gradient from both branches at
every step, an effective higher early learning rate. The freezing control is
consistent with it but does not isolate it; a direct test would rescale the
learning rate of the separate variant. The lesson generalizes: a single-run
result on a tiny memorization task is a hypothesis until it survives a
step-budget control, a data-scale control and, where a mechanism is
suspected, a control that isolates it.

\subsection{Curriculum and memory gate}

Table~\ref{tab:stages} splits one 600-step run (baseline, seed 0) by
curriculum stage (200 steps each). Losses are not comparable across stages
because the objective changes with $(p_{\max},\alpha)$; the higher tail mean
of the transition stage is consistent with the reweighting ramp raising the
loss scale. This is a single seed, and no run without the curriculum exists.

\begin{table}[t]
\centering
\small
\begin{tabular}{lrr}
\toprule
Stage & First-step loss & Tail-mean loss \\
\midrule
MAE ($p_{\max}=0.3$, $\alpha=0$) & 8.663 & 1.485 \\
Transition (linear ramp) & 1.419 & 2.007 \\
Diffusion ($p_{\max}=1.0$, $\alpha=1$) & 1.873 & 1.555 \\
\bottomrule
\end{tabular}
\caption{Per-stage loss of a single run
(\nolinkurl{docs/experiments/planned_experiments.md}, item 13); the tail
definition is given there.}
\label{tab:stages}
\end{table}

The memory gate was tested through the full training loop for 3 epochs (12
optimizer steps, so absolute losses are not comparable to the rest of the
study): min.\ loss $7.512$ without the gate, $7.453$ with the previous batch's
state only, $7.423$ with a random bank, and final loss $12.927$, $12.893$ and
$9.752$. Twelve steps cannot separate the variants; we draw no conclusion.

\section{Discussion}

\paragraph{Summary of findings.} The tiny networks reach a near-zero loss on the
memorization task, so the implementation optimizes; this does not show that the model
learns transferable structure. Taken together, the observations suggest that:
\begin{itemize}
  \item the codec ranking of Table~\ref{tab:seeds} is stable across three seeds, but it was
        measured after 600 steps on one repeated sentence and is a memorization-speed ranking;
  \item the hybrid attention (\texttt{liz2}) reaches a lower loss than window attention alone
        at 600 steps on every seed, and both reach the same floor at 3\,000 steps;
  \item the shared-projection advantage is a convergence-speed effect (\S\ref{sec:qkv-sharing});
        the mechanism is consistent with the freezing control but not isolated;
  \item depth, top-1 versus top-2 routing and the AttnRes block size show no measurable
        effect in this setup, the block-size result being a saturation effect;
  \item the curriculum and the memory gate have no supporting result: one seed for the
        former, twelve optimizer steps for the latter.
\end{itemize}

\paragraph{Practical implications.} The ablation runs used a CPU. A memorization test of this
kind can be repeated whenever a component changes, but the seed spread ($\pm0.002$ to
$\pm0.193$ in Table~\ref{tab:seeds}) limits which differences it can detect.

\paragraph{Compute sizing of the target.}
\label{sec:sizing}
The 200M-total/25M-active design point has not been instantiated; a roofline
argument~\citep{williams2009roofline} indicates what a run would need. Training costs
$6N_{\text{act}}D=4.5\times10^{18}$ FLOPs for $D=30$B tokens~\citep{kaplan2020scaling}, a ratio
$D/N_{\text{act}}=1{,}200$ far below that of the dense Pythia-14M, trained on about 300B
tokens~\citep{biderman2023pythia}. Chinchilla is fitted on dense models of 70M parameters and
more~\citep{hoffmann2022chinchilla}, so applying it to 25M active parameters is an extrapolation,
and MoE scaling laws~\citep{ludziejewski2025moe,abnar2025parameters} parameterize loss rather
than step time. Under balanced routing, a microbatch of $T$ tokens touches nearly every expert
once $T$ exceeds a few times $E/k$, so a step reads $N_{\text{total}}$ weights, not
$N_{\text{act}}$; with bf16 weights the arithmetic intensity is
$\mathcal{I}(T)=(N_{\text{act}}/N_{\text{total}})\,T$ FLOP/byte, and the smallest microbatch that
reaches the ridge point $\mathcal{I}^{\ast}=\pi/b_s$ of a GPU with peak throughput $\pi$ and
bandwidth $b_s$ is
\begin{equation}
T_{\min}=\frac{N_{\text{total}}}{N_{\text{act}}}\,\mathcal{I}^{\ast}.
\label{eq:tmin}
\end{equation}

With the vendor figures of a T4, an A100 and an H100~\citep{nvidia2018t4,nvidia2020a100,nvidia2022h100}
and $N_{\text{total}}/N_{\text{act}}=8$, $T_{\min}$ is about 1.6k tokens (T4, A100) to 2.4k tokens
(H100), against 200--300 for a dense model of the same size; below it, throughput is set by
$N_{\text{total}}$, not by the active parameters that FLOP-centric accounting counts. Sparsity is
therefore not free on memory-limited hardware. These are upper bounds from a model that counts
weight traffic only and omits activations, DeltaNet states, codec convolutions, MoE dispatch and
communication, and any utilization would be assumed, not measured. Whether a 16~GB card holds a
2.4k-token microbatch of this architecture is an empirical question this report does not answer.
The numbers come from \texttt{scripts/roofline.py}.

\subsection{Limitations}

The ablation study memorizes one sentence in 16 copies on text only, with a window branch that
attends globally, on CPU, with runs that are not bit-reproducible; its metric is a minimum over
noisy per-step losses. Only Table~\ref{tab:seeds} and \S\ref{sec:qkv-sharing} carry multi-seed
or multi-control statistics; the other sweeps are single-run. Parameter counts of the ablation
rows in Table~\ref{tab:seeds} other than the baseline are carried over from earlier logs. The
notebook config's active-parameter count has not been logged from an instantiated model. No dense baseline
is compared, so nothing here supports a claim of superiority; the multimodal, curriculum and
memory-gate components have no supporting result yet. The target sizing is analytical.

\subsection{Future work}
\label{sec:future}

The long-term aim behind Kairos is a small multimodal foundation model that a single local
machine can pretrain end to end; this report only asks whether the substrate optimizes. The
next step toward that aim is a pretraining run, which would turn these sanity checks into
held-out results. The pretraining notebook (\texttt{notebook/kairos\_pretraining.py}) trains on the text corpus
\texttt{keep-it-simple}~\citep{keepitsimple2026} (640{,}646 rows of bidirectional
prompt--text pairs) and on \texttt{keep-it-simple-multimodal}~\citep{keepitsimplemm2026}
(6{,}336 rows, a deliberately small pilot mix), through the
MAE, transition and diffusion curriculum. Planned measurements: (i) held-out loss per curriculum
stage and per modality against trivial baselines, and text loss with and without the
multimodal data; (ii) a comparison with Pythia-14M~\citep{biderman2023pythia} in bits per byte,
since tokenizers and objectives differ. Equal total parameter counts do not mean equal compute:
Pythia's parameters are largely embeddings of a vocabulary of about 50k entries, whereas a
byte-level vocabulary puts nearly all Kairos parameters in the backbone, so total,
non-embedding and active parameters and FLOPs per byte should all be reported. A Pythia-shaped
model trained on the same data would separate architecture from data, and a held-out corpus
that neither model has seen would avoid favoring Kairos; (iii) 3-seed, steps-to-threshold
reruns of the single-run sweeps on diverse data; (iv) a run without the curriculum; (v) a
measured tokens/s against microbatch size for the 200M/25M and a 50M/25M variant, which tests
Eq.~\ref{eq:tmin} directly.

\section{Conclusion}

Kairos poses a design question at 1--15M parameters: whether a hybrid bidirectional
DeltaNet/window-attention, MoE, denoising architecture over a byte-level space optimizes at
all. Memorization tests answer yes at this scale. The clearest architectural effect in these tests, a
shared-projection advantage, was a convergence-speed artifact, which is why we recommend
steps-to-threshold and control runs before reading a tiny-scale comparison. Generalization,
multimodal training and comparison with existing models are the objectives of the run described
in \S\ref{sec:future}; the 200M/25M design point of \S\ref{sec:sizing} is a sizing exercise only.

\section*{Code and reproducibility}

Code, configurations and logs are in the repository. Ablation configurations
are in \texttt{configs/ablations/}; a run is
\texttt{kairos overfit configs/ablations/<name>.yaml --steps 600 --mask-p-max 0.3
--no-mask-reweight --seed <0|1|2>}. The sizing numbers of \S\ref{sec:sizing} come from
\texttt{scripts/roofline.py}, checked by \texttt{tests/test\_roofline.py} and
\texttt{tests/test\_paper.py}. The pretraining notebook (marimo) is
\texttt{notebook/kairos\_pretraining.py}.

\bibliographystyle{plainnat}
\bibliography{kairos_references}

\end{document}